# DataWords: Getting Contrarian with Text, Structured Data and Explanations

**Stephen I. Gallant** sgallant@textician.com
**Mirza Nasir Hossain** nasir.hossain@textician.com

**Textician**

Cambridge, MA 02138
November 8, 2021

## Abstract

Our goal is to build classification models using a combination of free-text and structured data. To do this, we represent structured data by text sentences, *DataWords*, so that similar data items are mapped into the same sentence. This permits modeling a mixture of text and structured data by using only text-modeling algorithms. Several examples illustrate that it is possible to improve text classification performance by first running extraction tools (named entity recognition), then converting the output to DataWords, and adding the DataWords to the original text -- before model building and classification. This approach also allows us to produce explanations for inferences in terms of both free text and structured data.

## 1 Introduction

We seek to build predictive models using *free-text* in conjunction with *structured data*, while being able to explain (justify) model responses. For example, with healthcare modeling we need to take advantage of medical free-text (progress notes, lab reports), as well as database data (blood pressure, pulse, current medications). Applications include predicting ICD-10 codes as part of a Computer Assisted Coding (CAC) system, and warning of impending conditions such as sepsis [Gallant b].

When modeling text and structured data, there are several commonly accepted views ("conventional wisdom") that we question:

1. It is easier to build models from structured data than to build models from free-text.

2. The best way to build a model using both free-text and structured data is to first build two separate models, and then form a combined model using a (weighted) sum of the predictions for each model.

We challenge both points in the following sections, and present a convenient way to model a combination of free-text and structured data inputs, while being able to provide justifications for model inferences. Surprisingly, *our method employs free-text modeling only*, without using structured data modeling.

We give several examples predicting ICD-9 and ICD-10 medical codes. This paper does not claim state-of-the-art performance on any standard tests; it gives a simple new method for building models from a combination of free-text and structured data. We saw improved results in our tests predicting ICD codes.

## 2 Structured data is harder to model than free text

Structured data is usually considered easier to model than modeling free-text, because we can employ traditional statistical approaches such as regression, as well as neural network/deep learning approaches. Although the algorithmic modeling steps may be easier with structured data, the practical model-building process *as a whole* is often quite labor-intensive, and can be very difficult. This is particularly true of medical applications involving the Electronic Health Record (EHR). Structured data requires much effort prior to applying statistical or machine learning algorithms, including:

- Variable selection: Multiple expert-level meetings are needed to devise lists of important variables to include in modeling. Using all – thousands – of variables from the Electronic Health Record is not practical. For applications that involve thousands of models, such as ICD-10 Computer-Assisted Coding (CAC), variable selection becomes problematic, because different models (diabetes, flu, …) would benefit from their own different subsets of variables.
- Database field identification: A particular variable might reside in many similarly-named EHR database fields, requiring help from experts in the current EHR architecture to determine which precise database fields to incorporate into a model.
- Missing data: Always requires attention and effort with traditional structured data modeling.
- Data preparation: There is not a single blood pressure measurement – there may be dozens entered at various times. This requires appropriate rolling up of the numerical data (mean, median, max, min, temporal changes, …).

These are just some of the difficulties involved with modeling structured data from the EHR.

Turning to *unstructured data*, modeling free-text can also be challenging. Subtleties of language (e.g., similarities between different words or phrases, negation, and ambiguous word sense) can be very important. These and other subtleties make it hard to build good predictive models from text and, truth be told, building fully human-level computational models remains an unsolved problem. Nevertheless, there has been much recent progress building and automating text modeling, primarily using vector representations of words and text where everything is represented by lists of numbers. Examples include word2vec [Mikolov a,b], GloVe [Pennington], BERT [Devlin], MBAT [Gallant a], and older work with Salton's vector space model [Salton], Latent Semantic Indexing [Deerwester], and Context Vectors [Caid].

When modeling text using these methods, note that the actual model-building employs, behind the scenes, structured data algorithms such as regression. Of course structured data modelling issues must still be addressed. Perhaps the biggest one is missing data – most words only appear in a very few documents. Various vector-based free text modeling methods use different techniques for this issue, such as weighting schemes (tf-idf), singular value decomposition, and high dimensional vectors. We will not go into text modeling details here, other than to note that vector-based free text modeling methods use "good enough" structured data algorithms at their core so that modeling issues do not overwhelm them.

Perhaps the most widespread approach to unstructured text involves creating rules, such as: "*if the text includes 'diagnose cancer' then add 'cancer' to the list of diagnoses*". A problem with rules-based approaches is that they can require an enormous number of manually-created rules which interact with each other. For example, with Computer-Assisted Medical Coding software for ICD-10 codes, some commercial rules-based approaches require over 500,000 manually-coded rules! Rule interactions lead to "brittleness", and rules require significant effort to maintain. Shifting to a different coding scheme (CPT, SNOMED, the coming ICD-11) requires a massive amount of work. Using rules, it is also impractical to accommodate local phrasing, abbreviations, and formatting conventions that are particular to a single health organization. Such localization of modeling would require difficult rule set edits. By employing text-modeling approaches, we eliminate many of the disadvantages of rule-based approaches: all that is required is a good set of labeled examples. However, on the positive side for rules, performance may be very good for language constructs that are successfully captured by a rule set.

The current hot area in machine learning involves deep learning [Socher; LeCun] and convolutional neural networks [Kalchbrenner]. These approaches can produce top-performing models in terms of predictive accuracy, but they require a great amount of data and computational power for model building. They are harder to fully automate, and are also problematic with respect to extracting explanations of inferences, a subject receiving increased attention as "explainable AI" [Holzinger].

Comparing modeling from unstructured text vs. modeling from structured data, an important feature of unstructured EHR text (such as a "progress note") is that it explicitly mentions many structured data items, but *only if they are relevant to the case at hand*. For example, a physician will mention a patient's temperature only if it enters into a diagnosis or treatment (either positively or negatively). Similarly, she will mention specific medical conditions or drugs, but only if relevant to diagnosis or treatment. Thus, medical free-text contains a valuable resource: a highly-curated subsample of the structured data.

This motivates extracting structured data from medical free-text to help improve modeling performance by increasing the focus on these items. Then the question is what to do with the extracted values? Simply throwing them into a database for traditional analysis would be ill-advised. For example, there would be a major problem handling missing data, because most EHR variables are not mentioned in the text for an individual encounter. Here we will describe a way to leverage structured data extracts taken from unstructured text by adding special "sentences" to the original text, and thereby better "focusing" machine learning approaches onto this important subset of the text.

Logically, extracting structured data from free text adds no additional information to what is already in the free text. If we had perfect free-text modeling algorithms – or even human-level capabilities – such extraction would be a wasted effort. However, free-text machine learning is far from perfect. In studies (see below) we saw that adding structured data extracts of items that clinicians mention can be made to improve performance of these algorithms by increasing their focus on these aspects.

Of course, we can also obtain structured data directly from the EHR database, which has the advantage of providing values that are not included in the clinician notes. However, this aspect can also be a drawback if it brings in irrelevant values, i.e. noise. It may also require an extra data selection step, to produce an appropriate subset of variables for different patient encounters. We further examine these tradeoffs in experiments below.

## 3 Directly combining free-text and structured data models

Assume we are given both free-text and structured-data models for predicting the same target, where the target might be ICD-10 code "Q24.6" (Congenital heart block). It is certainly straightforward to create a combined model using a linear combination of the two predicted probabilities:

$w_1$ * <free-text prediction> + $w_2$ * <structured data prediction>.

If we are not taking an equally weighted combination, then this requires an additional modeling step to find the ($w_1$, $w_2$) parameters for the combination. Also, giving justifications for inferences from the combined model is not straightforward, as we need to somehow combine justifications from both the free-text and structured data models.

Recently [Zhang] has employed two different deep learning methods to combine text and structured data for three important hospital risk predictions. All weights are generated in the course of modeling.

We propose a different approach.

# 4 Modeling structured data by treating it as text

For this work, our starting point is the ability to model and make predictions from unstructured text. This is currently done in our software product, referred to as “NoNLP” because it avoids rules-based processing.[1] Our modeling software also identifies *key sentences* that justify the classification of a piece of text, by producing predictive scores for each sentence and taking the highest scoring sentences.

It is important to emphasize that text methods have the advantage of avoiding the cumbersome data preparation required for working with structured data: variable selection, numerical data preparation, and missing data. Could we beneficially use *only text methods* with the combination of text and structured data? This is a core innovation we explore in this paper.

Our approach to improving medical text modeling, particularly where a separate database is not available, is to first apply extraction methods to EHR free text, which produces only those structured data values that clinicians thought worthy of mentioning in their narratives. Then we turn this structured data into a type of free text that we can use with *existing* text modeling methods. Thus our (admittedly contrarian) approach *gives a way to use text methods to process a combination of text and structured/numeric data, where everything is expressed as text*.

We name our approach “DataWords”.

# 5 DataWords

The DataWords approach turns each item of extracted structured data, for example “**Temp = 98.8**”, into a *single sentence*, such as “**dw__Temp__mid_range.**”. (This example involves a one-word sentence.)

With respect to turning structured data into text, we want to have “**Temp = 98.3**” to be the same as “**Temp = 98.4**”. If we treat them as raw text, the strings are different – and for text, this means totally different. Thus, a requirement for treating structured data as text is representing *very similar* pieces of data as the *same* text.

Similarly, “**Temp = 101.0**” and “**Temp = 103.8**” are both high temperatures, which may be a key modeling differentiator for some targets. However, the latter is a significantly higher temperature, and this may well be a differentiator when modeling some other targets.

Thus, “**Temp = 102.1**” becomes the sentence “**dw__Temp__high_range.**”.

For “**Temp = 104.3**”, the sentence contains two words, “**dw__Temp__high_range dw__Temp__very_high_range.**” (Using two DataWords to represent very-high temperature helps machine learning recognize cases where temperature might be either high or very-high.)

Similarly, DataWords sentences can be created for low and very low temperatures.

In general, a single DataWord has a *data name* part, e.g. “**_Temp_**”, and a *data value* part, e.g. “**_high_range_**”. The data value part can also consist of textual values, e.g. “**_negative_**”, when associated with tests, such as “**_lime_disease_test_**”.

As mentioned, a key objective with DataWords transformations is to turn *similar* structured data values into the *same* DataWords, thereby helping the modeling learn more quickly and produce more robust results.

Each item of structured data will get its own sentence. This means that when generating explanations, text machinery can pick out a key sentence as justification, and that sentence may be either a DataWords sentence (referring to a structured variable) or a regular sentence referring to the free-text.

[1] See Textician.com for additional details and an online demo.

When reporting justifications in a user interface, we can also present a DataWords sentence to the user using a more natural sentence, such as "Temperature was very high [104.3]".

Some important points:

- Most text modeling methods automatically weight repeated words using, for example, tf-idf weighting [Salton]. They also handle *any* word, even newly created words as we are using here.
- Diagnoses, current medications lists, and other categorical structured data may be handled similarly, for example "**dw__Previous_condition__lung_cancer.**".
- Clinicians can specify their own thresholds for determining (mid, high, very_high, etc.) ranges for individual numeric variables. Alternatively, we can do a preliminary pass over training data to compute means and standard deviations for numeric data, and then use these statistics to define range automatically. For example, the high range can be defined to be

  **> (mean + 1.7 standard deviations**).

## 6 Explanations

There is increasing emphasis on being able to "explain" or "justify" predicted inferences. Computer Assisted Coding (CAC) gives a good example of the importance of such explanations. The goal of CAC is automated software to help medical coders generate appropriate disease codes, such as ICD-10 codes, from medical free text, such as "progress notes". These codes drive medical reimbursements from payers, and can require justifications from the original free text, such as supporting sentences or lab values. In Textician's CAC system, as applied to free-text, this is fairly simple to achieve. We merely apply the predictive model for the disease code in question to each sentence in the text, and return one or more key sentences as justification for calling that code.

A key benefit of using DataWords sentences to encode structured data is that *we can simply treat DataWords sentences exactly like normal free-text sentences for generating explanations*. As before, we simply generate key sentences for a particular ICD10 code model, and output resulting sentences, which may be either normal text sentences or DataWords sentences. If desired, it is easy to filter only text sentences or only DataWords sentences. Thus, we can make use of text explanation capabilities for free-text to include DataWords explanations, without making any major changes.

## 7 Processing Summary

To summarize processing using named entity recognition and DataWords, we combine three technologies: extraction capabilities, DataWords representations, and text modeling capabilities. This gives a way to hypothetically improve modeling from just the free text, as well as giving a means to combine both healthcare text and extracted structured/numeric information for building models and for making predictions. Moreover, we are able to justify inferences by pointing to key sentences and items of structured information.

Our overall approach for combining text and extracted structured/numeric data consists of the following processing steps:

1. Apply extraction to unstructured text to yield key structured data from the text (or take structured data from an associated database).
2. Represent each piece of structured data as a single DataWords sentence.
3. Take the original text and add all newly created DataWords sentences.
4. Use text modeling software on the combined text to build predictive models.
5. To classify new text, first apply steps 1-3 on the new text. Then apply the models created in Step 4 to the resulting combination of text + DataWords sentences.
6. Report predictions and justification sentences, where justifications consist of either sentences from the original text, or DataWords sentences representing a structured variable's value.

# 8 Case Studies

## 8.1 Using extraction

As noted, we can obtain structured data either from using extraction software on the original text, or by accessing a database that is associated with the text. We have experimented with the following extraction software:

- **ScispaCy** [Neumann] is a python package containing multiple models pre-trained on biomedical, scientific and clinical text. ScispaCy is built on spaCy, an open-source advanced natural language processing library, and enables Named Entity Extraction (NER), POS tagging, dependency parsing and word vectors, among other features. ScispaCy extracts entities using task-specific biomedical models (for example, extracting disease or chemical entities use a different pre-trained model versus gene or protein entities), and then maps them to a UMLS space via a KB-linker. ScispaCy models are neural networks trained on large corpus and comprising millions of parameters. Adding scispaCy to our software stack slowed down performance by 15x while processing clinical notes. We anticipate that the performance would be much better on GPU systems.
- **Comprehend Medical** [Bhatia] is a NER & Relationship Extraction (RE) service launched under Amazon Web Services and trained using state-of-the-art deep learning models. Being a web-service, Comprehend Medical does not require extensive training, installation, or pipeline configurations. It can also detect medical conditions and link them to ICD-10-CM codes. Being a cloud service presents a problem when personal health information is involved. It also has text size limitations that are smaller than many medical progress notes. It is possible to combine multiple API calls to process large medical notes, but this results in significant usage costs. For these reasons, we were only able to test extraction performance on a reduced subset of data with shorter text lengths.
- **Stanza** [Qi] is an open-source Python natural language processing toolkit supporting 66 human languages. Stanza features a language-agnostic, fully neural pipeline for text analysis, tokenization, multi-word token expansion, lemmatization, part-of-speech tagging, morphological feature-tagging, dependency parsing and named entity recognition (NER). Stanza is built with highly accurate neural network components, with modules built on top of the PyTorch library. Thus, it gets must faster performance on GPUs and in our experience the performance on CPUs is non-competitive with other open-source extractors.
- **CliNER** [Boag] is an easy to install (and use) open-source tool for extracting concepts (problems, tests and treatments) from clinical notes. CliNER 2.0 uses a word and character level bi-directional LSTM model and achieves competitive performance using pre-trained models. The neural network components cause considerable latency issues on CPU and our guess is that, like Stanza, it will provide greater performance on GPU configurations.
- **cTakes** [Savova] is perhaps the oldest NLP system that extracts clinical information from electronic health records. It processes clinical notes, identifying types of clinical named entities — drugs, diseases/disorders, signs/symptoms, anatomical sites and procedures. Each named entity has attributes for the text span, the ontology mapping code, context (family history of, current, unrelated to patient), and negated/not negated. cTakes is written in Java and is deployed using the Tomcat web-service. In our tests it beat the other extractors on sheer extraction performance, but the runtime performance was prohibitive. We did not evaluate cTakes on a GPU instance to validate increase in performance, but our analysis is that single-threaded Tomcat service was the primary bottleneck. An optimum configuration may bring down runtime performance, which would make cTakes the clear winner in the pre-processing NER race for our follow-on application.

**Datasets**

For our experiments we used two independent and proprietary Computer Assisted Coding data (ICD-

10), as well as MIMIC-III (ICD-9) [Johnson]. MIMIC-III ('Medical Information Mart for Intensive Care') is a large, single-center database comprising information relating to patients admitted to critical care units at a large tertiary care hospital.

We used a subset of 1148 examples from MIMIC-III. This was randomly selected from the full MIMIC datasets consisting of 58,362 clinical notes. We used a subset because most named entity extractors are much slower than our model-building software, some taking up to 12 hours to complete a single cross validation fold using this small subset. While this subset is a small fraction of the entire corpus, we've seen that it is a good representation of the full MIMIC-III corpus and in separate tests (not involving extraction) we have observed comparable performance on the subset and the full set with model-building. For our subset of MIMIC encounters, the median and average number of space-delimited terms in the texts were 21,792 and 43,554 respectively. MIMIC texts are anonymized progress notes, and are full of abbreviations, misspellings, local usages, etc. Extracts were quite noisy. Modeling targets were ICD-9 codes in MIMIC.

All tests are 4-fold CV.

### 1. Proprietary "S" Dataset

# encounters: 748
# documents: 2107
# codes: 32829
# Unique ICD-10 codes: 1901
Avg Codes per document: 15.58

| Dataset | DataWords | NER | F1 | Precision | Recall | Time per fold (mins) |
|---|---|---|---|---|---|---|
| **Proprietary "S"** | No | NA | 0.357 | 0.461 | 0.342 | ~5 |
| **Proprietary "S"** | Yes | ScispaCy | 0.388 | 0.491 | 0.367 | ~18 |
| **Proprietary "S"** | Yes | Stanza | 0.371 | 0.470 | 0.352 | ~22 |

### 2. Proprietary "P" Dataset

# encounters: 966
# documents: 966
# Unique ICD-10 codes: 1749

| Dataset | DataWords | NER | F1 | Precision | Recall |
|---|---|---|---|---|---|
| **Proprietary "P"** | No | NA | .29 | .22 | .25 |
| **Proprietary "P"** | Yes | ScispaCy | .36 | .28 | .31 |

**3. Mimic-III (ICD-9)** (subset)

# encounters: 1148
# documents: 1148
# codes: 9867
# Unique codes: 1571
Avg Codes per document: 8.59

| DataWords | NER | F1 | Precision | Recall | Time per fold (mins) |
|---|---|---|---|---|---|
| **No** | NA | 0.340 | 0.504 | 0.322 | ~5 |
| **Yes** | ScispaCy | 0.349 | 0.500 | 0.334 | ~65 |
| **Yes** | Stanza | 0.344 | 0.521 | 0.324 | ~125 |
| **Yes** | CliNER | 0.345 | 0.508 | 0.325 | ~550 |
| **Yes** | cTAKES | 0.365 | 0.495 | 0.351 | ~720 |

**4. Mimic-III (ICD-9)** (different subset with shorter text lengths)

| DataWords | NER | F1 | Precision | Recall | Comment |
|---|---|---|---|---|---|
| **No** | NA | 0.355 | 0.614 | 0.324 | |
| **Yes** | AWS Comprehend | 0.358 | 0.607 | 0.330 | |
| **Yes** | AWS Comprehend | 0.367 | 0.544 | 0.342 | Just DataSentences; no text |
| **Yes** | AWS Comprehend | 0.375 | 0.558 | 0.357 | Just non-numeric DataSentences; no text |

For modeling, we used proprietary Textician software, which constructs a fixed-length vector representation for each document, employs a type of regression to model each code, and a final threshold fitting step. Software is very fast for model-building and for scoring.

## 8.2 Using structured database values

We also experimented with using structured data values taken directly from the MIMIC database. For this experiment we used a different MIMIC subsample consisting of 2000 clinical notes, sampled randomly from the complete corpus. To reduce noise from value types that appeared very infrequently, we limited ourselves to those values that occurred at least 100 times. Some types of data had many entries, for example Glucose readings appeared thousands of times. We used various subsamples of measurement types, as summarized below.

1. All measurements irrespective of their count.
2. Measurements that have a minimum of 100 and a maximum of 500 count.
3. Top 100 measurements (by count).

4. Top 500 measurements (by count).
5. Top 100 measurements after excluding the original top 100 (3)

When creating numerical values into DataWords sentences, we used the following numerical thresholds as defaults:

> **Very low**: less than (mean – 1.7 standard deviations), else
> **Low**: less than (mean – 1.0 standard deviations), else
> **Mid**: less than (mean + 1.0 standard deviations), else
> **High**: less than (mean + 1.7 standard deviations), else
> **Very high**.

Of course, it is possible to specify other thresholds for size boundaries, as well as to have a different number of groupings by size.

Performance improvement using structured data directly was less than when using extraction from the text for these experiments.

| Subsample type | F1 | Precision | Recall |
|---|---|---|---|
| Baseline (no structured data) | 0.329 | 0.417 | 0.322 |
| All measurements irrespective of their count | 0.327 | 0.445 | 0.304 |
| Measurements with count <=100 & >= 500 | 0.332 | 0.433 | 0.317 |
| Top 100 measurements by count | 0.325 | 0.430 | 0.309 |
| Top 500 measurements by count | 0.326 | 0.437 | 0.306 |
| Top 100 measurements after excluding original top 100 | 0.331 | 0.431 | 0.316 |

We have performed previous research predicting severe sepsis using separate free-text and structured data models, but well before developing DataWords technology [Culliton]. We were able to get improved performance by combining outputs from two models versus using just a free-text model. However, predicting a single target, severe sepsis, makes it easier for experts to select, by committee, which structured data values to use. Committee selections would not be viable when predicting thousands of different targets, as with ICD-10 codes.

## 9 Discussion

A big advantage of extracting from the original text is that Clinicians usually limit their mentions of structured data values to those values that are important for this particular case. This gives a valuable, curated subset of structured data values that makes use of Clinician expertise.

The disadvantage of using structured data extracts from the original text is that extraction software makes many errors, especially when applied to notes produced by Clinicians having abbreviations, typos, local jargon, etc. Using extraction software is also slow, taking 15 to 150 times as long as just building the text models!

By contrast, taking values directly from a database produces many more values that are not relevant to the case at hand, unless we first perform a time-consuming analysis to select database variables. Moreover, an extra variable reduction step by experts is especially impractical when many different types of cases are being examined, such as is the case with Computer Assisted Coding applications. Such irrelevant database values function as noise, and they can overwhelm relevant values and the text itself.

It is interesting to speculate whether the brain uses a distributed representation analogous to DataWords as one of its representation methods.

In conclusion, we summarize our claims and findings:

1. Contrary to prevailing wisdom, it is often easier to build free-text models than to build structured data models (provided you have decent text modeling software).
2. Using DataWords techniques, we can convert structured data to text, permitting us to use text modeling methods on the combination.
3. Applying extraction software to free-text, followed by DataWords conversions, tends to result in better performance than modeling just the free text. We have found ScispaCy to improve follow-on DataWords modeling performance (3% to 25%), with acceptable processing time delay.
4. If we extract structured data *directly from a database* and combine it with text, it appears harder to achieve improved model performance.